%% file: Paper-SWoMo-Arxiv.tex
\documentclass[runningheads]{llncs}
\usepackage[T1]{fontenc}
\usepackage{graphicx,verbatim}
\usepackage{booktabs}
\usepackage{float}
\usepackage{makecell}
\usepackage{pifont}
\usepackage{amsmath} 
\usepackage{amssymb}
\usepackage{array}

\usepackage[table]{xcolor}
\usepackage{xcolor}
\usepackage{colortbl}
\definecolor{1gray}{HTML}{e2efda} 
\definecolor{2gray}{HTML}{a9d08e}      
\definecolor{3gray}{HTML}{7b9c65}  
\definecolor{green}{HTML}{228B22}
\definecolor{red}{HTML}{E60000}

\usepackage[hidelinks]{hyperref}

\begin{document}

\title{SWoMo: Neuro-Symbolic World Model for Cataract Surgery Simulation}

\author{Ssharvien Kumar Sivakumar\inst{1,2}\orcidID{0009-0009-0383-9462}
Akwele Johnson\inst{1} \and
Anirudh Dhingra\inst{1} \and
Yannik Frisch\inst{1,3} \and
Ghazal Ghazaei\inst{2} \and
Anirban Mukhopadhyay\inst{1}}


\institute{Technical University Darmstadt, Darmstadt, Germany \email{ssharvien\_kumar.sivakumar@tu-darmstadt.de}
 \and Carl Zeiss AG, Munich, Germany
 \and AICM, Medical Faculty of Heidelberg University, Heidelberg University, Germany}

\maketitle              
\begin{abstract}
Realistic surgical simulation plays a crucial role in training novice surgeons and in the development of autonomous agents. World models can scale such simulation environments to realistic and diverse procedures by predicting future patient states conditioned on current observations and surgical actions. However, current state-of-the-art approaches often fail to satisfy key criteria required for clinical applicability, including visual realism, physically grounded interactions, and the ability to simulate scenarios beyond the training distribution. Hence, we introduce SWoMo, a neuro-symbolic world model for cataract surgery simulation that decouples motion generation from visual realism. The symbolic component, consisting of a rule-based simulator and scene graph representations, models motion dynamics and tool-tissue interactions, while a diffusion model produces realistic visual appearance, including textures and tissue deformations. We propose an inverse pairing strategy that reconstructs real surgical videos in the simulator to obtain paired simulated and real videos, which are then used to train our video diffusion model for the reverse objective of sim-to-real translation. Our experiments show both qualitative and quantitative improvements over prior work. We demonstrate that our simulator further satisfies the key criteria, including generalisation to unseen interaction geometries, improvements in downstream phase detection, and unsupervised video style transfer. The code, data, and model weights are available at: \url{https://ssharvienkumar.github.io/SWoMo/}.


\end{abstract}

\input{sections_arxiv/introduction}
\input{sections_arxiv/method}
\input{sections_arxiv/experiment}
\input{sections_arxiv/conclusion}





%
%
%

\bibliographystyle{splncs04}
\bibliography{Paper-SWoMo}
%




\end{document}

%% file: sections_arxiv/introduction.tex
\section{Introduction}

Surgery demands high precision, where irreversible actions must be carefully coordinated through structured spatial reasoning and causal understanding~\cite{chen2025far}. Simulated environments provide a safe and controllable setting in which such dependencies can be explored \cite{frisch2025surgrid}, enabling applications ranging from risk-free, immersive surgical skills training for novice surgeons to the development of autonomous surgical robotic agents that can reason and plan interactively \cite{lin2025visuomotor,nair2021effectiveness}. However, scaling such environments to realistic and diverse procedures requires models that integrate continuous perception with explicit representations of the geometry and dynamics of the surgical scene \cite{boels2025surgical}. Surgical world models address this need by enabling \textit{interactive simulation} and \textit{predictive modelling} of future patient states conditioned on current observations and actions \cite{ha2018world,koju2025surgical}, thereby \textit{unifying perception, dynamics, and action} within a single framework \cite{boels2025surgical}.

However, the clinical applicability of a surgical world model hinges on satisfying three key criteria: \textbf{(I)} The simulated environment must achieve \textbf{high visual realism} to minimise the sim-to-real gap, as low fidelity limits policy transfer~\cite{kadian2020sim2real}. Yet this remains a major challenge for traditional simulators, particularly in modelling complex instrument-tissue interactions \cite{koju2025surgical,sivakumar2025sg2vid}. \textbf{(II)} Equally important is ensuring \textbf{physically grounded interactions}, so that both learning agents and human trainees acquire behaviours that transfer reliably to real high-stakes surgical settings \cite{yang2025physworld}. \textbf{(III)} The \textbf{dynamic representation of the world model must go beyond the training distribution}, enabling the generation of plausible outcomes under unseen tool geometries, insertion angles, and novel surgical workflows.

Unfortunately, state-of-the-art research typically excels on one criterion while failing to satisfy others. Methods that attempt to control video generative models via various conditioning signals \cite{biagini2025hierasurg,he2022latent,niu2024mofa,sivakumar2025sg2vid} fall short of interactive surgical simulation due to the lack of step-wise action conditioning. As a result, they provide limited fine-grained control for agent or trainee intervention and cannot be considered true world models. Their behaviour is also tightly coupled to the training distribution, degrading significantly under out-of-distribution (OOD) conditions~\cite{chen2025far}. World-model based approaches, such as SurgWM~\cite{koju2025surgical}, inspired by Genie \cite{bruce2024genie}, learn interactive environments from unstructured surgical videos, but limit interaction to latent action codes rather than direct tool control. DreamGen~\cite{jang2025dreamgen} and SurgWorld \cite{he2025surgworld} generate paired video-action data using inverse dynamics, improving policy learning, yet they lack physical grounding. Unpaired image translation methods \cite{martyniak2025simuscope,venkatesh2024surgical} have also been explored to address the sim-to-real gap, but remain insufficient for modelling complex spatiotemporal tool-tissue interactions.

\begin{figure}
\includegraphics[width=\textwidth]{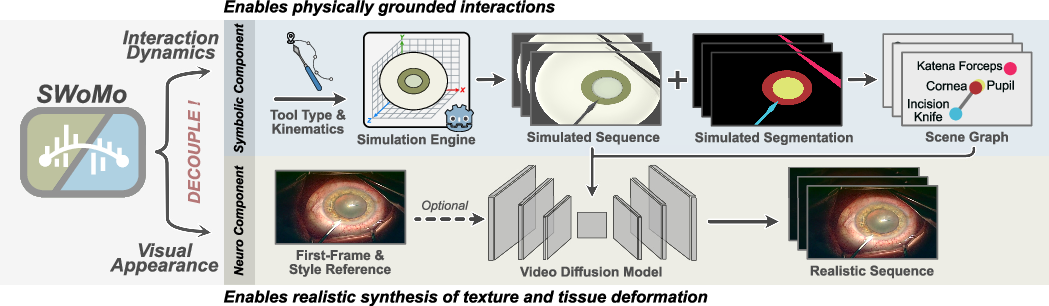}
\caption{\textbf{Neuro-symbolic World Model} for interactive cataract surgery simulation that decouples surgical interaction dynamics from visual appearance.} 
\label{fig:concept_figure}
\end{figure}

Hence, we introduce SWoMo, an interactive neuro-symbolic world model for cataract surgery, as shown in Figure~\ref{fig:concept_figure}, in which \textbf{symbolic scene graphs and a rule-based physics simulator} explicitly encode structure and constraints, while a \textbf{diffusion model} provides high-fidelity visual synthesis. Our proposed method \textit{explicitly decouples the modelling of surgical motion and interaction dynamics from visual appearance}. This allows precise tool and anatomical motions to be preserved through physically grounded simulation while enabling the generation of entirely novel scenarios beyond the training distribution. Visual realism, such as texture and tissue deformation, is learned through data-driven synthesis using a diffusion model. To achieve this, we introduce \textbf{inverse pairing strategy} in which tool and anatomical motions are extracted from real surgical videos and replayed in the simulator, producing large-scale paired simulated and real videos. These pairs are then used to train a video-to-video diffusion model that translates simulated renderings back into realistic surgical videos. The simulator also produces corresponding segmentations, which are used to construct scene graphs that efficiently encode the scene and object relationships in a structured graphical format. We show that conditioning the diffusion model on scene graphs is crucial for mitigating issues caused by residual misalignment between simulated and real videos. Optional initial frame or style reference conditioning is added for subject-specific surgical simulation. We further demonstrate generalisation to unseen surgical interactions, notable improvements in downstream phase detection, and unsupervised video style transfer across datasets.

%% file: sections_arxiv/method.tex
\section{Method}

Our surgical world model decouples motion generation from visual realism. Motion and interaction dynamics are governed by symbolic components comprising of scene graphs $\mathcal{G}$ and a rule-based simulator $T(\cdot)$, while visual appearance is generated by a denoising diffusion model, employing a neural network $\boldsymbol{\epsilon}_\theta(\cdot)$. We begin by extracting the anatomical configuration $k^{\text{anat}}_t$ and tool kinematics $k^{\text{tool}}_t$ from real surgical video $x_{1:T}$, and use them to construct a digital twin of the eye and replay these signals in this simulator. At time step $t$, we define the simulator state as $\bar{x}_t = \{{k^{\text{anat}}_t, k^{\text{tool}}_t}\}$. Actions correspond to explicit surgical tool motions, $a_t = \Delta k^{\text{tool}}_t$. State transitions are governed by $T(\cdot)$ such that $(\bar{x}_{t+1}, \bar{m}_{t+1}) = T(\bar{x}_t, a_t)$, where $\bar{x}_{t+1}$ denotes the next state along with its rendered simulated frame, and $\bar{m}_{t+1}$ is the corresponding segmentation. From each simulator state, a scene graph is generated, $\mathcal{G}_t = S(\bar{x}_t, \bar{m}_t),$ encoding object geometry and relational structure. Final observations are then generated according to $p_\theta(x_{t:t+n}\mid \bar{x}_{t:t+n}, \mathcal{G}_{t:t+n})$, parameterized by the diffusion model, which translates simulated sequence $\bar{x}_{t:t+n}$ into realistic sequence $x_{t:t+n}$. Here, $n$ denotes the number of consecutive simulated frames jointly provided as conditioning to the diffusion model.

\textbf{Inverse Pairing:}
In this section, we describe the extraction of $k^{\text{anat}}_t$ and $k^{\text{tool}}_t$ from real surgical video $x_{1:T}$. We start by segmenting pupil and iris employing nnU-Net \cite{isensee2021nnu}, and corresponding segmentations are fitted with ellipses to obtain compact geometric parameters $k^{\text{iris}}_t$ and $k^{\text{pupil}}_t$ encoding centroid, orientation, and axis lengths. Motion of the eye globe decomposes into global and local components. \textit{Global motion}, arising from camera or patient movement, is estimated by annotating skin landmarks in the first frame $x_{1}$ and tracking them over time using pre-trained CoTracker \cite{karaev2023cotracker}. The resulting trajectories define a global transformation $g_t$ used to estimate the displacement of the eye globe centroid. \textit{Local motion} corresponds to rotational movement of the eye globe and is derived from the centroid trajectory of the pupil mask, yielding rotational parameters $r_t$. Formally, the globe motion is represented as $k^{\text{globe}}_t = \{g_t, r_t\}$. The anatomical configuration is thus defined as $k^{\text{anat}}_t = \{{k^{\text{globe}}_t, k^{\text{iris}}_t, k^{\text{pupil}}_t}\}$. Tool kinematics are recovered from tool masks generated using SASVi \cite{sivakumar2025sasvi} and SAM2-based \cite{ravi2024sam} manual interactive annotation tool. From these masks, we extract geometric parameters $k^{\text{tool}}_t = \{{c_t, \theta_t,\beta_t}\}$ where $c_t$ denotes tool tip position, $\theta_t$ orientation, and $\beta_t$ articulation parameters such as bending angle and opening angle.

\textbf{Rule-based Simulator:}
The recovered anatomical and tool parameters $\{k^{\text{anat}}_t, k^{\text{tool}}_t\}$ are used to drive a rule-based simulator $T(\cdot)$ on Godot game engine \cite{godot_engine}. The simulator instantiates a parameterised eye globe and surgical tools, whose mesh transformations are directly controlled by these parameters. The global translation $g_t$ is mapped from image coordinates to simulator space through normalisation and fixed scaling, defining the eye globe position, while rotational parameters $r_t$ control eye globe orientation via yaw and pitch. Anatomical meshes are updated using $k^{\text{iris}}_t$ and $k^{\text{pupil}}_t$. Iris and pupil geometry are scaled according to their estimated axis lengths, enabling subject-specific variation.

Tool meshes are controlled by $k^{\text{tool}}_t = \{c_t, \theta_t, \beta_t\}$. Tool tip positions $c_t$ are mapped to simulator space using the same normalisation as global globe motion, while orientations $\theta_t$ directly determine tool rotation. For articulated instruments such as forceps, articulation parameters $\beta_t$ control relative mesh rotations to reproduce opening motions, while for angled tools they adjust the shaft bend to maintain geometric consistency. Final tool positions are defined relative to the eye globe surface with an offset scaled by the globe’s anatomical scaling factor, ensuring consistent placement in the simulated video $\bar{x}_{1:T}$. We present a real video and its simulated pair, along with segmentations, in Supplementary~E.

\textbf{Symbolic Scene Graph}:
We form scene graphs $\mathcal{G}_{1:n}$ from $\bar{x}_{1:n}$ and its segmentation $\bar{m}_{1:n}$ and subsequently encode $\mathcal{G}_{1:n}$ following a strategy inspired by SG2VID \cite{sivakumar2025sg2vid}. Each node in $\mathcal{G}_{1:n}$ represents a connected component from $\bar{m}_{1:n}$ and stores high-level component attributes, including centroid, spatial spreading, and average optical flow within the corresponding region. We pre-trained two separate graph encoders, global encoder $\mathit{E}_{\mathcal{G}}^{glob}$ and local encoder $\mathit{E}_{\mathcal{G}}^{loc}$. The $\mathit{E}_{\mathcal{G}}^{glob}$ learns high-level structural relationships via contrastive learning between $\mathcal{G}_{1:n}$ and $\bar{m}_{1:n}$. In contrast, $\mathit{E}_{\mathcal{G}}^{loc}$ focuses on appearance cues by learning to reconstruct masked regions in the real sequence $x_{{1:n}}$ using information from $\mathcal{G}_{1:n}$. We find the additional $\mathcal{G}_{1:n}$ conditioning to be important for visual fidelity and conditional adherence, as also highlighted by the ablation study in Table~\ref{tab:quantitative_result}. It contributes in two main ways: \textit{first}, $\mathcal{G}_{1:n}$ provides explicit class-level information that is not available from the $\bar{x}_{1:n}$ alone, helping the model distinguish object identities. \textit{Second}, we observe regions near component boundaries, especially near tool-tissue interaction, often suffer from degraded synthesis due to misalignments between $\bar{x}_{1:n}$ used for video conditioning and $x_{{1:n}}$, which introduce conflicting training signals. When such inconsistencies occur repeatedly, the model tends to average visual features across boundaries, causing smoothing artifacts. Incorporating $\mathcal{G}_{1:n}$ alleviates this by abstracting local pixel-level misalignments, as the node's component-level attributes remain relatively stable under small boundary shifts. Moreover, $\mathcal{G}_{1:n}$ provides only higher-level structural guidance, allowing generator to infer realistic visual boundaries by itself from the learned appearance priors rather than strictly following misaligned $\bar{x}_{1:n}$ conditioning. 

\begin{figure}
\includegraphics[width=\textwidth]{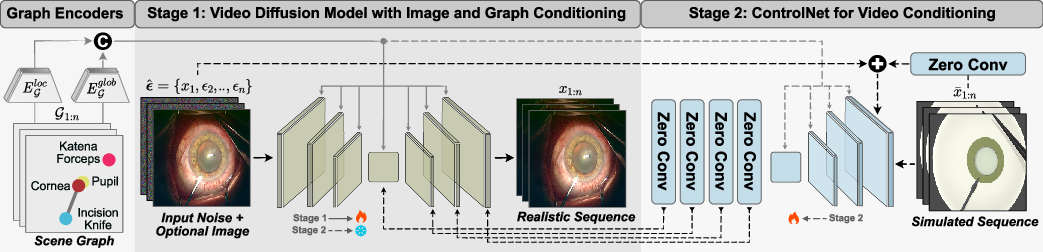}
\caption{\textbf{Overview of SWoMo’s} two-stage video diffusion training.}
\label{fig:architecture}
\end{figure}

\textbf{Video Diffusion Model with Graph-Image-Video Conditioning}:
Our diffusion model for translating simulated sequence $\bar{x}_{t:t+n}$ into realistic sequence $x_{t:t+n}$ is trained in two stages, as illustrated in Figure~\ref{fig:architecture}: \textit{first}, learning image and graph to video generation, and \textit{subsequently} incorporating video-level conditioning through ControlNet \cite{zhang2023adding} training. Diffusion models \cite{ho2020denoising} rely on a parameterised network $\boldsymbol{\epsilon}_\theta$ that is trained to reverse the gradual noise injection process $p(x_{t,\tau-1} \mid x_{t,\tau}, c)$, where $\tau$ denotes the diffusion timestep. Training is performed by minimizing a mean squared error objective between the true noise and the noise predicted by $\boldsymbol{\epsilon}_\theta$: $\min_{\theta}\mathbb{E}_{\tau, x_{t,0}, \epsilon} \left[ \left\| \epsilon - \epsilon_\theta(x_{t,\tau}, \tau, c) \right\|^2 \right]$. To extend diffusion models to video generation \cite{ho2022video}, the model is augmented with temporal layers. Specifically, temporal convolution and attention layers are interleaved with spatial layers, enabling the model to capture temporal dependencies and motion dynamics across frames. In the first stage, the conditioning signal $c$ consists of the first-frame $x_{{1}}$ and the sequence scene graphs $\mathcal{G}_{1:n}$. For first-frame conditioning, the noise term $\epsilon_1$ is replaced with the actual first-frame $x_{{1}}$. The resulting model input is therefore constructed as $\hat{\boldsymbol{\epsilon}} = \{x_{{1}}, \epsilon_2, \epsilon_3, \dots, \epsilon_n \}$. For graph conditioning, we concatenate the outputs of $\mathit{E}_{\mathcal{G}}^{glob}$ and $\mathit{E}_{\mathcal{G}}^{loc}$ and further concatenate the result with the timestep embedding before passing it to $\boldsymbol{\epsilon}_\theta$.

In the \textit{second stage}, we enable conditioning on the simulated sequence $\bar{x}_{1:n}$. For that, we freeze the parameters $\theta$ of the pre-trained diffusion backbone $\boldsymbol{\epsilon}_\theta$ from the previous stage and create a separate, trainable copy of its encoder with parameters $\theta_c$. The frozen backbone $\boldsymbol{\epsilon}_\theta$ and the trainable encoder are connected through zero-initialised convolutional layers at multiple resolutions \cite{zhang2023adding}, allowing the conditioning simulated sequence $\bar{x}_{1:n}$ to modulate intermediate feature representations without disrupting the pretrained generative prior of $\boldsymbol{\epsilon}_\theta$ and preserving its visual quality. The $\bar{x}_{1:n}$ is encoded through this control branch, and the resulting features are injected into the corresponding layers of $\boldsymbol{\epsilon}_\theta$. 

%% file: sections_arxiv/experiment.tex
\section{Experiments and Results}

We evaluate our method on two publicly available cataract surgery datasets: CATARACTS \cite{al2019cataracts} and Cataract-1k \cite{ghamsarian2023cataract}. To limit the need for extensive frame-level annotations, we restrict our modelling to safety-critical phases, which together account for around half of the video: \textit{Idle, Incision, Viscoelastic, Capsulorhexis, Hydrodissection}, and \textit{Phacoemulsification}. After this filtering, the datasets contain 868 videos from Cataract-1k and 50 videos from CATARACTS, with a mean duration of just over three minutes. For training, videos are temporally sampled at 4 frames per second into sequences of 16 frames and spatially resized to 128×128. Data splits are created at the video level, using a 50/6/44 ratio for CATARACTS and 80/10/10 for Cataract-1k. The training of the diffusion model is distributed across four NVIDIA A40 GPUs.

\begin{table}[h!]
    \centering
    \caption{Quantitative Comparisons of Synthesis Quality and Conditioning Adherence.}
    \fontsize{8}{8}\selectfont
    \label{tab:quantitative_result}
    \begin{tabular}{lccccc ccccc}
        \midrule
        & \multicolumn{5}{c}{\textbf{CATARACTS} \cite{al2019cataracts}} & \multicolumn{5}{c}{\textbf{Cataract-1k} \cite{ghamsarian2023cataract}}  \\
        \cmidrule(lr){2-6} \cmidrule(lr){7-11}
        Method & FVD$\downarrow$ & FID$\downarrow$ & LPIPS$\uparrow$ & BB IoU$\uparrow$ & F1$\uparrow$ & FVD$\downarrow$ & FID$\downarrow$ & LPIPS$\uparrow$ & BB IoU$\uparrow$ & F1$\uparrow$ \\
        \midrule
        StyleGAN-V \cite{skorokhodov2022stylegan} & 581.5 & 107.2 & 0.379 & -- & -- & 501.2 & 116.4 & 0.286 & -- & -- \\
        Endora \cite{li2024endora} & 436.8 & 58.4 & 0.456 & -- & -- & 258.7 & 40.0 & 0.379 & -- & -- \\
        MedSora \cite{wang2024optical} & 1243.3 & 127.6 & 0.403 & -- & -- & 809.9 & 147.6 & 0.324 & -- & -- \\
        LVDM \cite{he2022latent} & 1604.6 & 131.0 & \cellcolor{3gray}0.557 & 0.228 & 0.154 & 1469.6 & 176.6 & \cellcolor{3gray}0.519 & 0.213 & 0.188 \\
        MOFA \cite{niu2024mofa} & 993.4 & 105.6 & 0.446 & 0.432 & 0.282 & 716.7 & 94.6 & 0.358 & 0.418 & 0.404 \\
        SG2VID \cite{sivakumar2025sg2vid} & \cellcolor{1gray}363.8 & \cellcolor{1gray}47.3 & 0.436 & \cellcolor{1gray}0.497 & \cellcolor{1gray}0.391 & \cellcolor{3gray}73.0 & \cellcolor{3gray}14.9 & \cellcolor{1gray}0.392 & \cellcolor{1gray}0.607 & \cellcolor{1gray}0.623 \\
        \midrule
        SWoMo & \cellcolor{3gray}265.4 & \cellcolor{3gray}40.8 & 0.450 & \cellcolor{3gray}0.522 & \cellcolor{3gray}0.412 & \cellcolor{2gray}123.0 & \cellcolor{2gray}20.1 & 0.388 & \cellcolor{3gray}0.645 & \cellcolor{3gray}0.656 \\
        SWoMo-\ding{55}IMG  & \cellcolor{2gray}329.3 & \cellcolor{2gray}42.3 & \cellcolor{1gray}0.451 & \cellcolor{2gray}0.514 & \cellcolor{2gray}0.406 & \cellcolor{1gray}134.3 & \cellcolor{1gray}20.2 & 0.389 & \cellcolor{2gray}0.622 & \cellcolor{2gray}0.642 \\
        SWoMo-\ding{55}SG & 390.7 & 52.1 & \cellcolor{2gray}0.463 & 0.377 & 0.296 & 283.6 & 35.5 & \cellcolor{2gray}0.434 & 0.529 & 0.554 \\
        \midrule
    \end{tabular}
\end{table}

\textbf{Quantitative Comparison and Ablation:} We evaluate the quality and diversity of synthesised videos using FID, FVD \cite{unterthiner2018fvd}, and the LPIPS diversity score \cite{zhang2018lpips}, as shown in Table~\ref{tab:quantitative_result}. For conditional methods, we additionally measure adherence to conditioning using detection-based metrics, including F1 score and bounding box IoU. Specifically, we train Mask2Former \cite{cheng2021masked} to detect tools and anatomical structures, and compare bounding box predictions across synthesised and real videos. Our method is benchmarked against multiple baselines with publicly available implementations. StyleGAN-V \cite{skorokhodov2022stylegan}, Endora \cite{li2024endora}, and MedSora \cite{wang2024optical} represent unconditional video generation approaches. We further modify LVDM \cite{he2022latent} to enable text-based conditioning using scene graph triplets, whereas SG2VID \cite{sivakumar2025sg2vid} directly conditions on scene graphs. MOFA \cite{niu2024mofa} conditions generation on both the initial frame and motion trajectory derived from sparse optical flow. SWoMo outperforms most baselines in terms of visual fidelity and achieves substantial improvements over all baselines in terms of conditioning adherence. In Table~\ref{tab:quantitative_result}, we also provide ablations without scene graph conditioning (SWoMo-\ding{55}SG) and without image conditioning (SWoMo-\ding{55}IMG) to illustrate the contribution of each conditioning component.

\textbf{Qualitative Assessment:} We present SWoMo's qualitative results in Figure~\ref{fig:qualitative}, with additional samples in Supplementary A and a qualitative comparison in Supplementary B. SWoMo leverages both the simulated video and an initial frame for video synthesis, whereas SWoMo-\ding{55}IMG relies solely on the simulated video. These results highlight how SWoMo’s sim-to-real transfer task reduces major implementation effort on the simulator, for example, by eliminating the need to explicitly model textures and deformation, while still accurately following tool movements and anatomical configurations from the simulated video.

\begin{figure}
\includegraphics[width=\textwidth]{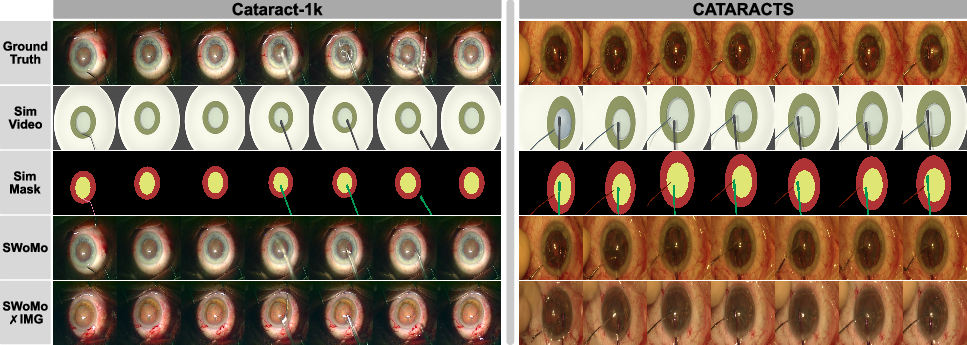}
\caption{\textbf{Qualitative Results of Sim-to-Real Video Transfer.}} 
\label{fig:qualitative}
\end{figure}

\textbf{Unsupervised Video Style Transfer:} 
SWoMo enables unsupervised style transfer between the CATARACTS and Cataract-1k domains without training on paired sequences. This is achieved through a shared intermediate representation of the simulated sequence that is used across both datasets. Specifically, we utilise the simulated sequence from the source domain with the initial-frame conditioning from the target domain to synthesise videos in the target style. Because the intermediate representation encodes geometry and motion independently of appearance, the model preserves interaction dynamics while adopting the visual characteristics of the target domain. The style transfer results are shown in Figure~\ref{fig:style_ood} (top) for both datasets, with full videos provided in Supplementary~C.

\textbf{Improved Generalisation to Novel Tool Motions:} Generative models are fundamentally constrained by their training distributions and often degrade on sequences far outside these distributions. By introducing an intermediate simulated sequence representation that provides explicit structural and motion guidance, we further push the boundaries of what the model can handle. Using the simulator, we generate difficult cases, including tools with novel entry directions and tool combinations that never co-occur in the training data. Visual results are shown in Figure~\ref{fig:style_ood} (bottom), with additional examples provided in Supplementary D, demonstrating that the intermediate representation substantially improves generalisation under OOD conditions.

\begin{figure}
\includegraphics[width=\textwidth]{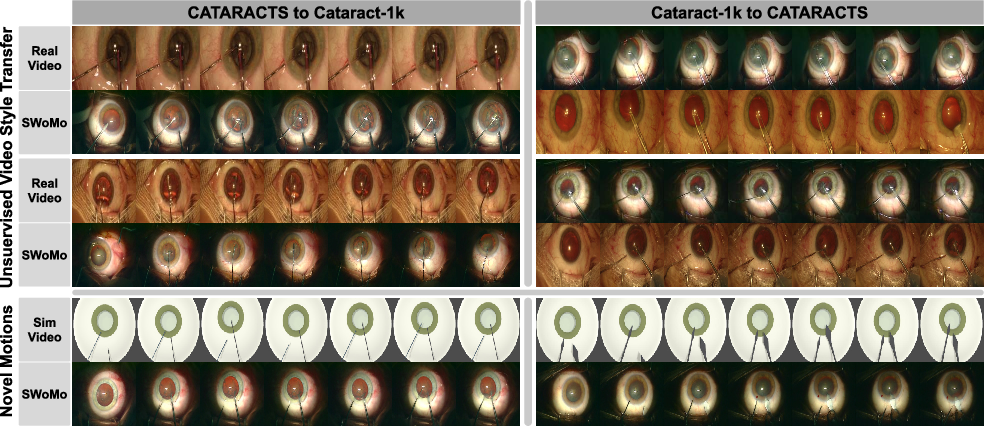}
\caption{Unsupervised Video Style Transfer and Generalisation to Novel Tool Motion} 
\label{fig:style_ood}
\end{figure}

\textbf{Downstream Evaluation on Phase Recognition:} We use synthesised videos to augment the training data of a downstream model for phase recognition during cataract surgery. The phase recognition is performed using MS-TCN++ \cite{li2020ms} trained on DINO features \cite{caron2021emerging}. To generate a synthesised video, two real videos are randomly selected from the training set. The visual style is taken from the first video, while the tool and anatomical motion patterns are taken from the second. Phase annotations are inherited from the second video, since the underlying surgical actions remain unchanged, making this process a form of generative data augmentation. To generate full-length surgical videos, sequences are synthesised autoregressively by first generating a sequence and then using its last frame as the first-frame conditioning input for the subsequent sequence. Using this strategy, we effectively double the number of training videos. Table~\ref{tab:downstream_results} presents the results, demonstrating improvements over training on real data alone and also over generative augmentation with other methods.

\begin{table}[h]
    \centering
    \caption{Performance on Downstream Phase Recognition.}
    \fontsize{8}{9}\selectfont
    \label{tab:downstream_results}
    \begin{tabular}{lcc cc}
        \midrule
        & \multicolumn{2}{c}{\textbf{CATARACTS} \cite{al2019cataracts}}  
        & \multicolumn{2}{c}{\textbf{Cataract-1k} \cite{ghamsarian2023cataract}} \\
        \cmidrule(lr){2-3} \cmidrule(lr){4-5}
        Training data & Accuracy$\uparrow$ & F1-Score$\uparrow$ & Accuracy$\uparrow$ & F1-Score$\uparrow$ \\
        \midrule
        Real Only & 79.4 & 79.3 & 93.7 & 94.9 \\
        \midrule
        Real + LVDM \cite{he2022latent} & 63.6 ({\color{red}{-15.8}}) & 65.3 ({\color{red}{-14.0}}) & 71.3 ({\color{red}{-22.4}}) & 72.4 ({\color{red}{–22.5}}) \\
        Real + SG2VID \cite{sivakumar2025sg2vid} & 80.5 ({\color{green}{+1.1}}) & 81.6 ({\color{green}{+2.3}}) & 94.0 ({\color{green}{+0.3}}) & 95.3 ({\color{green}{+0.4}}) \\
        Real + SWoMo (Ours) & 82.5 ({\color{green}{+3.1}}) & 83.0 ({\color{green}{+3.7}}) & 94.2 ({\color{green}{+0.5}}) & 95.2 ({\color{green}{+0.3}}) \\
        \midrule
    \end{tabular}
\end{table}

%% file: sections_arxiv/conclusion.tex
\section{Conclusion}
We present SWoMo, tool kinematics are fed into a rule-based simulator, whose outputs are then converted into realistic videos, effectively combining the strengths of physical simulation and generative modelling. We demonstrate several unique capabilities of SWoMo, including precise unsupervised video style transfer without additional training and strong generalisation to unseen interaction geometries with novel tool motions and co-occurrences. Finally, we show that the generated videos can effectively augment existing datasets and improve performance on downstream tasks such as surgical phase recognition.